\documentclass[a4paper]{article}
\usepackage{INTERSPEECH2020}
\usepackage{url}
\usepackage{hyperref}
\usepackage{graphicx}
\usepackage{amssymb,amsmath,bm}
\usepackage{algorithm}
\usepackage{algpseudocode}
\usepackage{textcomp}
\usepackage{comment}
\usepackage{multirow}

\def\vec#1{\ensuremath{\bm{{#1}}}}

\usepackage{amsmath,amsfonts,bm}

\def\eqref#1{equation~\ref{#1}}
\def\1{\bm{1}}

\def\vtheta{{\bm{\theta}}}

\def\vx{{\bm{x}}}
\def\vy{{\bm{y}}}

\DeclareMathAlphabet{\mathsfit}{\encodingdefault}{\sfdefault}{m}{sl}
\SetMathAlphabet{\mathsfit}{bold}{\encodingdefault}{\sfdefault}{bx}{n}

\title{An Investigation of Few-Shot Learning in Spoken Term Classification}
\usepackage{blindtext}
\makeatletter
\def\name#1{\gdef\@name{#1\\}}
\makeatother
\name{{\textit{ Yangbin Chen$^{1}$, Tom Ko$^{2\dag}$, Lifeng Shang$^3$, Xiao Chen$^3$, Xin Jiang$^3$, Qing Li$^4$} %\thanks{\dag This work is done when the first author worked as intern at Huawei Noah's Ark Lab.} 
\thanks{\dag corresponding author.}}}
\address{$^1$Department of Computer Science, City University of Hong Kong \\ $^2$Department of Computer Science and Engineering, \\ Southern University of Science and Technology, Shenzhen, China \\ $^3$Huawei Noah's Ark Lab \\ $^4$Department of Computing, The Hong Kong Polytechnic University}
\email{robinchen2-c@my.cityu.edu.hk, tomkocse@gmail.com, \\  \{shang.lifeng,chen.xiao2,jiang.xin\}@huawei.com,csqli@comp.polyu.edu.hk}
\begin{document}
\maketitle
\begin{abstract}
In this paper, we investigate the feasibility of applying few-shot learning algorithms to a speech task.
We formulate a user-defined scenario of spoken term classification as a few-shot learning problem.
In most few-shot learning studies, it is assumed that all the $N$ classes are new in a $N$-way problem.
We suggest that this assumption can be relaxed and define a $N$+$M$-way problem where $N$ and $M$ are the number of new classes and fixed classes respectively.
We propose a modification to the Model-Agnostic Meta-Learning (MAML) algorithm to solve the problem.
Experiments on the Google Speech Commands dataset show that our approach\footnote{Code is available at: \url{https://github.com/Codelegant92/STC-MAML-PyTorch}} outperforms the conventional supervised learning approach and the original MAML.
\end{abstract}

\section{Introduction}
\label{sec:intro}

In recent years, few-shot learning has drawn a lot of attention in the machine learning community.
It tries to tackle a very challenging task of which the model has to adapt to new tasks with very few labeled examples.
A lot of elegant solutions have been developed and the most popular solution right now uses meta-learning.
Meanwhile, most of the studies on few-shot learning are conducted on image tasks.
It is worth to investigate the feasibility of applying few-shot learning algorithms to speech tasks.

In spoken term classification, the target spoken terms are usually predefined and known in advance.
Given a sufficient amount of training data, conventional supervised learning could have solved the problem nicely \cite{audhkhasi2017end, rosenberg2017end}.
However, when it comes to a user-defined scenario, the system performance degrades considerably if the user selects rare words.
\cite{sharma2020adaptation} attributes the degradation to the lack of training data and addresses the problem by data generation with Text-To-Speech (TTS) techniques.
In most studies of user-defined scenario \cite{Sacchi2019OpenVocabularyKS, szoke2005comparison, trmal2017kaldi}, users can only define new keywords in the same language which matches the internal phoneme set.
%However, there are lack of studies about where user can define new spoken terms.
%Many novel model architectures have been developed and a large number of training data have been accumulated \cite{he2017streaming}.
%In spite of the overall improvement in recent DNN-based KWS model, \cite{sharma2020adaptation} points out that performance  deteriorates  considerablyif the selected keyword is a rare word
%However, when users want to define some new spoken terms such as terms from another language whose phoneme sequences (or subsequences) are rare in training data, it is difficult to achieve a satisfied result.
%To address the problem, there have been some data-level solutions.
%One solution is to collect more data for each new term.
%However, it costs more time and money.
%Another solution is to do data augmentation such as generating synthetic speech using Text-To-Speech (TTS) technology \cite{sharma2020adaptation, murthy2018effect}.

In this paper, we want to simulate a user-defined scenario where users can define new spoken terms in any languages by providing a few audio examples.
We formulate this problem as a few-shot learning problem and investigate the performance of state-of-the-art model-level few-shot learning solutions.

Meta-learning, also known as `learning to learn', aims to make quick adaptation to new tasks with only a few examples.
Recently many different meta-learning solutions have been proposed to solve the few-shot learning problems. 
These solutions differ in the form of learning a shared metric \cite{vinyals2016matching,snell2017prototypical,sung2018learning,ko2020prototypical}, a generic inference network \cite{santoro2016meta,mishra2018a}, a shared optimization algorithm \cite{munkhdalai2017meta,Ravi2017OptimizationAA}, or a shared initialization for the model parameters \cite{finn2017model,li2017meta,nichol2018first}.
In this paper, we adopt the Model-Agnostic Meta-Learning (MAML) approach \cite{finn2017model} because of the following reasons: 
\begin{itemize}
    \item It is a very general framework and can be easily applied on a new task.
    \item It is model-agnostic.
    \item It achieves state-of-the-art performance in existing few-shot learning tasks.
%    \item It performs gradient-based rapid adaptation to target tasks.
\end{itemize}
To the best of our knowledge, there is no prior work of applying MAML on similar speech tasks.

Few-shot learning is often defined as a $N$-way, $K$-shot problem where $N$ is the number of classes in the target task and $K$ is the number of examples of each class. In most previous studies, it is assumed that all the N classes are new.
However, in real-life applications, these classes are not necessary to be all new. For example, in spoken term classification, the silence and the unknown (words that not belong to any keywords) classes are known in prior. 
Thus, we further define a $N$+$M$-way, $K$-shot problem where $N$ and $M$ are the number of new classes and fixed classes respectively.
In this task, the model has to concurrently classify among new classes and fixed classes.
We propose a modification to the original MAML algorithm to solve this problem.

We conduct our experiment on Google Speech Commands dataset \cite{warden2018speech} to simulate a user-defined scenario in spoken term classification. 
%In our experiment, we verify the effectiveness of our approach in the problem of few-shot spoken term classification.
We compare our approach with two baseline approaches:
the conventional supervised learning approach and the original MAML approach.
%Experiments on the Google Speech Commands dataset \cite{warden2018speech} show that our approach performs the best.
Experimental results show that our extended-MAML leads to obvious improvement over the two baselines.

Here summarizes our contributions in this paper:
\begin{itemize}
    \item We investigate the performance of MAML, as one of the most popular few-shot learning solutions, on a speech task.
    \item We extend the original MAML to solve a more realistic $N$+$M$-way, $K$-shot problem.
    \item We investigate how much a user-defined spoken term classification system can get close to a predefined one.
\end{itemize}

The rest of the paper is organized as follows. In Section 2 we present the basic of MAML. In Section 3 we introduce our approach for the few-shot spoken term classification problem. In Section 4 we describe the details of our experiments. Section 5 is the conclusion and future work.

\section{Model-Agnostic Meta Learning (MAML)}
\label{sec:maml}

\subsection{The basic idea}
\label{subsec:intro-maml}
\begin{figure}[t]
\includegraphics[width=8cm]{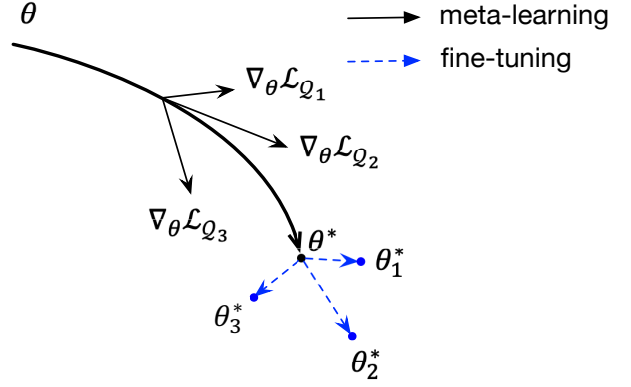}
\caption{The MAML algorithm learns a good parameter initializer $\vtheta^*$ by training across various meta-tasks such that it can adapt quickly to new tasks.}
\label{figure:1}
\end{figure}

MAML is one of the most popular meta-learning algorithms which aims to solve the few-shot learning problem. The goal of MAML is to train a model initializer which can adapt to any new task using only a few labeled examples and training iterations\cite{finn2017model}.
To reach this goal, the model is trained across a number of tasks and it treats the entire task as a training example.
The model is forced to face different tasks so that it can get used to adapting to new tasks.
In this section, we will describe the MAML training framework in a general manner. 
As is shown in Figure \ref{figure:1}, the optimization procedure consists of two stages.
We will first introduce the meta-learning stage on the training data then introduce the fine-tuning stage on the testing tasks.

\subsection{The meta-learning stage}
\label{subsec:maml-method}
Given that the target evaluation task is a $N$-way, $K$-shot task, the model is trained across a set of tasks $\mathcal{T}$ where each task $\mathcal{T}_i$ is also a $N$-way, $K$-shot task.
In each iteration, a learning task (a.k.a. meta-task) $\mathcal{T}_i$ is sampled according to a distribution over tasks $p(\mathcal{T})$. Each $\mathcal{T}_i$ consists of a support set $\mathcal{S}_i$ and a query set $\mathcal{Q}_i$.

Consider a model represented by a parametrized function $f_{\vtheta}$ with parameters ${\vtheta}$.
$\vtheta'_i$ is computed from $\vtheta$ through the adaptation to task $\mathcal{T}_i$. A loss function $\mathcal{L}_{\mathcal{S}_i}(f_{\vtheta})$, which is a cross-entropy loss over the support set examples, is defined to guide the computation of $\vtheta'_i$:
%First, the base model's parameter $\theta$ is updated to $\theta'_i$ through one or more gradient descent steps using the loss $\mathcal{L}_{\mathcal{S}_i}$.
%In our work, $\mathcal{L}_{\mathcal{S}_i}(f_{\theta})$ is a cross-entropy loss over the support set examples:

\begin{equation}\label{eq1}
\begin{split}
\mathcal{L}_{\mathcal{S}_i}(f_{\vtheta}) = -\sum_{(\vx_j,\vy_j) \in \mathcal{S}_i}{\vy_j logf_{\vtheta}(\vx_j)}
\end{split}
\end{equation}

A one-step gradient update is as below:

\begin{equation}\label{eq2}
\begin{split}
\vtheta_i' = \vtheta - \alpha\nabla_{\vtheta}\mathcal{L}_{\mathcal{S}_i}(f_{\vtheta})
\end{split}
\end{equation}
where $\alpha$ is the learning rate which can be a fixed hyperparameter or learned like the Meta-SGD \cite{li2017meta}. 
In practice, the gradient is often updated for several steps.

Then the model parameters are optimized on the performance of $f_{\vtheta'_i}$ evaluated by the query set $\mathcal{Q}_i$ with respect to ${\vtheta}$.
$\mathcal{L}_{\mathcal{Q}_i}(f_{\vtheta'_i})$ is another cross-entropy loss over the query set examples:

\begin{equation}\label{eq3}
\begin{split}
\mathcal{L}_{\mathcal{Q}_i}(f_{\vtheta'_i}) = -\sum_{(\vx'_u,\vy'_u) \in \mathcal{Q}_i}{\vy'_u logf_{\vtheta'}(\vx'_u)}
\end{split}
\end{equation}

Generally speaking, MAML aims to optimize the model parameters such that one or a small number of gradient steps on a new task will lead to maximally effective behavior on that task.
At the end of a training iteration, the parameters $\vtheta$ are updated as below:

\begin{equation}\label{eq4}
\begin{split}
\vtheta^* \leftarrow \vtheta - \beta\nabla_{\vtheta} \mathcal{L}_{\mathcal{Q}_i}(f_{\vtheta'_{i}})
%\theta \leftarrow \theta - \beta\nabla_{\theta} \sum_{\mathcal{T}_i \sim p(\mathcal{T})} \mathcal{L}_{\mathcal{T}_i}(f_{\theta'_{i}})
\end{split}
\end{equation}
where $\beta$ is the learning rate of the meta learner.
The loss computed from the query set results in a second-order\footnote{Please note that the second-order gradient optimization here is not equal to performing a first-order gradient optimization twice.} gradient optimization on $\vtheta$.

To increase the training stability, instead of a single task, usually a batch of tasks is sampled in each iteration.
The optimization is performed by averaging the loss across the tasks.
Thus, equation (\ref{eq4}) can be generalized to 

\begin{equation}\label{eq5}
\begin{split}
\vtheta^* \leftarrow \vtheta - \beta\nabla_{\vtheta} \sum_{i} \mathcal{L}_{\mathcal{Q}_i}(f_{\vtheta'_{i}})
\end{split}
\end{equation}

\subsection{The fine-tuning stage}
A fine-tuning is performed before the evaluation.
In a $N$-way, $K$-shot task, $K$ examples from each of the $N$ classes are available at this stage (the support set of the target task).
The model trained from the previous stage will be fine-tuned according to equation (\ref{eq2}) for a few iterations. 
Then the updated model will be evaluated on the remaining unlabeled examples (the query set of the target task).

\vspace{4mm}

\section{Few-shot spoken term classification}
\label{sec:fsscr}

\vspace{4mm}
\subsection{Motivation}
\label{subsec:motivation}
In section 2, it is assumed that all classes in the target task are new classes.
However, these classes are not necessary to be all new.
In real-life applications, some of the classes are known so that more examples of these classes can be used in the meta-learning stage.
In this paper, we call them fixed classes as we later fix their output positions in the neural network classifier.
We call this task, which has to concurrently classify among new classes and fixed classes, a $N$+$M$-way, $K$-shot problem where $N$, $M$, $K$ are the number of new classes, fixed classes and examples from each new class for fine-tuning respectively.
%This problem has not been investigated in the original work of MAML.
This problem of concurrently classifying unseen and seen classes has not been investigated in the original work of MAML.
In our work, we try to tackle the problem by proposing a modification to the MAML training framework. 
We believe that the $N$+$M$-way, $K$-shot problem is more realistic and our modification to MAML is applicable to a variety of different tasks.  
In this section, we will describe our methodology for a few-shot spoken term classification task.

\subsection{Our methodology}
\label{subsec:kws-negative}
Although the $N$+$M$-way, $K$-shot problem can be regarded as a specific form of the normal $N$-way, $K$-shot problem, solving it with the original MAML framework will lead to a degradation of performance.
By making use of the prior information of the $M$ fixed classes,
we modify the MAML framework in the following aspects:
\begin{itemize}
    \item We fix the output positions of the fixed classes in the neural network classifier.
    \item The fixed classes occur in every meta-task $\mathcal{T}_i$ in the meta-learning stage.
    \item The adaptation of fixed classes is not needed in the fine-tuning stage as they have already been learned in the meta-learning stage.
\end{itemize}
The above three extensions to the original MAML make the framework more effectively applied to real applications.

\begin{figure}[t]
\includegraphics[width=8cm]{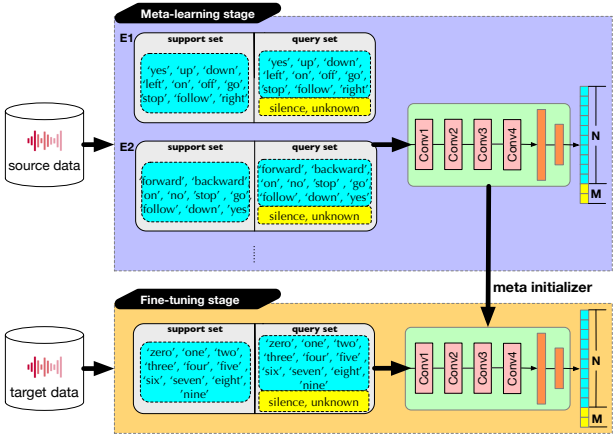}
\caption{Framework of our extended-MAML approach for few-shot spoken term classification.}
\label{Figure:2}
\end{figure}

\subsection{Spoken term classification}
\label{subsubsec:framework}
We formulate the user-defined spoken term classification task as a $N$+$M$-way, $K$-shot classification task.
$N$ is the number of keywords that users can define and users should define each keyword by providing $K$ audio examples.
$M$ is set to 2 in our work as we have two fixed classes: silence and unknown.
Here, the unknown class represents words that do not belong to any keywords.

Figure \ref{Figure:2} illustrates the framework of our extended-MAML approach.
The target data contain audio examples from $N$ user-defined keywords and two fixed classes, while the source data contain audio examples from totally different keywords except the two fixed classes.
In the meta-learning stage, a number of $N$+2-way, $K$-shot meta-tasks are sampled from source data.
Each meta-task consists of a support set and a query set.
The form of each meta-task is similar to the target task.
As we expect to learn a model initializer which can adapt to the target task using the user-defined keywords only, we exclude the fixed classes from the support set in both the meta-learning and the fine-tuning stages.
As we can assume availability of more training examples of the fixed classes, we keep them in the query set of all meta-tasks in the meta-learning stage.
Furthermore, it can be seen that the positions of the silence and the unknown classes are fixed to the last of the network output (the yellow area).
Thus, we force the model to ``remember" the fixed classes without the need of adaptation.
%Furthermore, as is shown in the grey areas in Figure \ref{Figure:1}, the query sets of both the training and test tasks collect examples from two fixed classes: silence and unknown.
%And the output layer of the base model assigns two fixed positions for the fixed classes.

Algorithm \ref{alg:maml} summarizes the details of our approach.
The algorithm is based on the work of \cite{finn2017model} but different in the sampling of the support set and the query set during the meta-training stage, which is introduced in Section 3.2.

\begin{algorithm}[t]
\caption{extended-MAML approach for few-shot spoken term classification}
\label{alg:maml}
\begin{algorithmic}[1]
    \Require
    $p(\mathcal{T}):$ distribution over tasks
    \Require
    $\mathcal{X}:$ training keywords set
    \Require
    $\mathcal{S}_{il}:$ silence class set, $\mathcal{U}_{nk}:$ unknown class set
    \Require
    $\mathcal{S}_i:$ support set, $\mathcal{Q}_i$: query set
    \Require
    $\alpha$, $\beta$: learning rates
\State Randomly initialize base model parameters $\vtheta$
\While{not done}
    \State Sample a batch of meta-tasks $\mathcal{T}_i \sim p(\mathcal{T})$
    \ForAll{$\mathcal{T}_i$}
        \State Sample a support set $\mathcal{S}_i$ from $\mathcal{X}$
        \State Compute the gradient $\nabla_{\vtheta}\mathcal{L}_{\mathcal{S}_i}(f_{\vtheta})$ using $\mathcal{S}_i$ and $\mathcal{L}_{\mathcal{S}_i}(f_{\vtheta})$ in Equation (\ref{eq1})
        \State Update base model parameters with gradient descent: $\vtheta'_i=\vtheta-\alpha\nabla_{\vtheta}\mathcal{L}_{\mathcal{S}_i}(f_{\vtheta})$ \Comment{step 6 and step 7 can be repeated for several times}
        \State Sample a query set $\mathcal{Q}_i$ from the union $\{\mathcal{X},\mathcal{S}_{il},\mathcal{U}_{nk}\}$  \Comment{selected keywords from $\mathcal{X}$ in $\mathcal{Q}_i$ and $\mathcal{S}_i$ within $\mathcal{T}_i$ are the same}
        \State Compute the loss $\mathcal{L}_{\mathcal{Q}_i}(f_{\theta'_i})$ using $\mathcal{Q}_i$ and the updated model $f_{\vtheta'}$ 
    \EndFor
    \State Update parameters $\vtheta$ using each $\mathcal{Q}_i$ and $\mathcal{L}_{\mathcal{Q}_i}(f_{\vtheta'})$:
     $\vtheta \leftarrow \vtheta - \beta \nabla_{\vtheta}\sum_{i} \mathcal{L}_{\mathcal{Q}_i}(f_{\vtheta'_i})$
\EndWhile
\end{algorithmic}
\end{algorithm}

%The details of our approach are described in Algorithm \ref{alg:maml}.
%During the training phase, we first sample a batch of tasks.
%For each task, we randomly select $N$ keywords from the training set to build up a support set.
%We sample datapoints from the keywords and use the cross-entropy loss over them to update the base model's parameters.
%Then we sample datapoints from the same $N$ keywords, together with the silence and unknown classes, to build up a query set.
%After the inner loop updates within the batch, the meta-learner of the outer loop will update the parameters using a second-order gradient optimization by averaging the cross-entropy losses over the query set datapoints across the tasks.

%During the testing phase, we get $N$ user-defined keywords.
%For each keyword, we have $K$ examples provided by the users to form the support set.
%We use them to fine-tune our initialized model trained from the previous stage and evaluate it on the query set which contains examples from the defined keywords as well as silence and unknown.

\section{Experiment}
\label{sec:exp}

\vspace{4mm}

\subsection{Experimental setup}

\subsubsection{Dataset}
We conduct our experiments on Google Speech Commands dataset (v0.02) \cite{warden2018speech}.
It consists of 105,829 1-second audio clips of 35 keywords.
We formulate two 10+2-way, $K$-shot tasks using the same setup as the “Audio Recognition” tutorial in the official Tensorflow package \cite{audiotutorial}.
The first task is digits classification, which uses digits zero to nine as ten user-defined keywords.
The second task is commands classification, which contains 10 user-defined keywords as: “yes”, “no”, “up”, “down”, “left”, “right”, “on”, “off”, “stop”, and “go”.
For each task, besides 10 user-defined keywords, we randomly pick 5 keywords to form the unknown class set and use the remaining 20 keywords to form the training keywords set.
We also generate audio examples of the silence class by mixing the background noise.
In the meta-learning stage, the training keywords set, unknown class set, and silence class set are used to form different meta-tasks $\mathcal{T}_i$.
The 10 user-defined keywords are unseen to the meta-learning stage and only $K$ labeled examples of each of them are available in the fine-tuning stage, where the initialized model is fine-tuned on the labeled examples and evaluated on the unlabeled examples.
%The keywords combined with `unknown’ and `silence’ form the training set.
%The digits combined with `unknown’ and `silence’ form the test set.

\subsubsection{Model Setting}
The 1-second clips are sampled at 16kHz.
We use Mel-Frequency Cepstral Coefficient (MFCC) features.
For each clip, we extract 40 dimensional MFCCs with a frame length of 30ms and a frame step of 10ms.
CNN is adopted as the base model which contains 4 convolutional blocks.
Each block comprises a 3 $\times$ 3 convolutions and 64 filters, followed by ReLU and batch normalization \cite{Ioffe2015BatchNA}.
The flattened layer after the convolutional blocks contains 576 neurons and is fully connected to the output layer with a linear function.
The models are trained with a mini-batch size of 16 for 1, 5, 10, 15, 20, 30-shot classification and 4 for 50, 100-shot classification.
We set the learning rate $\alpha$ to 0.1 and $\beta$ to 0.001.
%The base learner is trained with SGD optimizer with a learning rate $\alpha$ of 0.1 and the meta-learner is trained with Adam optimizer with a learning rate $\beta$ of 0.001.
%Then follows a low-rank linear layer with size 58 and a fully connected layer with size 128.
%The output layer is a softmax function.
%More details about the feature extraction and the base model can be found in \cite{zhang2017hello}.
%To make fair comparisons among different approaches, all experiments are 5-way-$K$-shot classification tasks.
%Each testing operation is repeated 100 times to weaken the influence of the randomness.
\subsubsection{Baselines}
We compare our proposed approach with two baseline approaches: the conventional supervised learning approach which trains the model on the support set of the target task only, and the original MAML which treats the 10+2-way problem as a 12-way problem. 
In the evaluation, we sample $K$ examples from each class for fine-tuning the model and 100 examples for evaluation. 
We do 100 times random tests and evaluate different approaches on accuracy.

\subsection{Results and discussions}
\subsubsection{Few-shot spoken term classification performance}
We compare our approach with two baselines.
Table \ref{tab:digits} and Table \ref{tab:commands} list the performance of digits classification and commands classification respectively on 1, 5, 10-shot tasks.
First of all, the overall accuracy on digits classification is better than that on commands classification\footnote{This observation is consistent with \cite{kao2019sub}.}. 
This implies that, in a user-defined spoken term classification, the system performance will be affected by the keywords users define.
Not surprisingly, MAML based approaches perform much better than conventional supervised learning in a few-shot situation.
Our proposed approach outperforms the original MAML.
We attribute the improvement to the use of prior information of the fixed classes.

\begin{table}[h]
    \centering
    \caption{Accuracy with 95\% confidence intervals on \textbf{digits classification}}
    \begin{tabular}{llll}
         \hline
 %        & \multicolumn{3}{c}{\bf 12-way} \\
        
         \multicolumn{1}{c}{\bf Methods} & \multicolumn{1}{c}{1-shot} & \multicolumn{1}{c}{5-shot} & \multicolumn{1}{c}{10-shot} \\
         \hline
         \hline
         Superv. L. & \multicolumn{1}{c}{18.14 $\pm$ 0.44} & \multicolumn{1}{c}{24.83 $\pm$ 0.38} & \multicolumn{1}{c}{28.07 $\pm$ 0.34} \\
         MAML-ori & \multicolumn{1}{c}{44.60 $\pm$ 0.98} & \multicolumn{1}{c}{60.88 $\pm$ 0.58} & \multicolumn{1}{c}{65.18 $\pm$ 0.62} \\
         MAML-ext & \multicolumn{1}{c}{\bf 47.42 $\pm$ 0.96} & \multicolumn{1}{c}{\bf 63.22 $\pm$ 0.71} & \multicolumn{1}{c}{\bf 69.48 $\pm$ 0.47} \\
         \hline
    \end{tabular}
    \label{tab:digits}
\end{table}

\begin{table}[h]
    \centering
    \caption{Accuracy with 95\% confidence intervals on \textbf{commands classification}}
    \begin{tabular}{llll}
         \hline
   %      & \multicolumn{3}{c}{\bf 12-way} \\
        
         \multicolumn{1}{c}{\bf Methods} & \multicolumn{1}{c}{1-shot} & \multicolumn{1}{c}{5-shot} & \multicolumn{1}{c}{10-shot} \\
         \hline
         \hline
         Superv. L. & \multicolumn{1}{c}{ 17.03 $\pm$ 0.48 } & \multicolumn{1}{c}{ 22.42 $\pm$ 0.33 } & \multicolumn{1}{c}{ 25.6 $\pm$ 0.26 } \\
         MAML-ori & \multicolumn{1}{c}{ 33.35 $\pm$ 0.80} & \multicolumn{1}{c}{ 50.31 $\pm$ 0.50} & \multicolumn{1}{c}{ 57.34 $\pm$ 0.41} \\
         MAML-ext & \multicolumn{1}{c}{\bf 39.54 $\pm$ 0.62} & \multicolumn{1}{c}{\bf 52.20 $\pm$ 0.51} & \multicolumn{1}{c}{\bf 59.36 $\pm$ 0.39} \\
         \hline
    \end{tabular}
    \label{tab:commands}
\end{table}

\subsubsection{User-defined vs. predefined}
\begin{figure}[t]
\includegraphics[width=8cm]{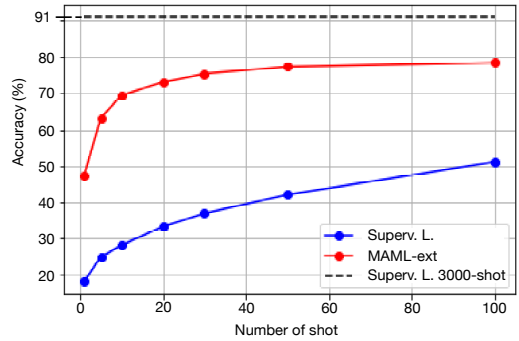}
\caption{Accuracy with changing shot on \textbf{digits classification}.}
\label{Figure:3}
\end{figure}

We take the result in \cite{kao2019sub} as a reference to the predefined scenario of the same task which has an average of about 3000 training examples per class.
%We take the performance of the supervised model trained on 3000-shot as a benchmark of the pre-defined scenario.
We further increase the number of shot in our approach to see if the performance of a few-shot system can get close to a predefined system.
%classical supervised learning with much training data.
Figure \ref{Figure:3} summarizes the results.
It can be seen that the performance of our approach (78.48\%) gets much closer to the 3000-shot system (91\%) than that of the conventional supervised model using few-shot.
However, there is still a performance gap between the two.
In the future, we will try to narrow the gap by incorporating more prior information to the meta-learning stage and applying data augmentation techniques \cite{ko2017study,tom2015speed,zhu2019mixup}.
%In the future, we will try to narrow the gap with data augmentation techniques \cite{ko2017study, tom2015speed}.

%As the 10-shot accuracy of our approach is lower than supervised model trained on a large quantity of examples like \cite{kao2019sub}, we try to figure out how the performance will be influenced by the number of shot so that our approach can get closer to real applications.
%We change the $K$ from 1 to 100 and draw a curve showing the accuracy change.
%From the figures we can see that it is possible for our approach to reach competitive results with no need of massive examples.

\section{Conclusions and Future work}
\label{sec:conclusion}
In this paper, we formulate a user-defined scenario of spoken term classification as a few-shot learning problem. 
We define it as a $N$+$M$-way, $K$-shot problem and propose a modification to the Model-Agnostic Meta Learning (MAML) algorithm to solve the problem.
%We address the imbalance problem and we incorporate the prior information of the negative class as external knowledge to our framework.
Experiments conducted on the Google Speech Commands dataset show that our approach performs the best compared to the baselines.
%For the experiments of few-shot recognition without the negative class, we find that the MAML outperforms the fundamental supervised learning and the feature transfer learning.
%For the experiments of few-shot recognition with the negative class, we find that incorporating the negative class to the meta-training process as external knowledge is effective.
We observe that there is a performance gap between a user-defined system and a predefined system.
In the future, we will try to narrow the gap with a combination of both the data augmentation techniques which are promising in improving model robustness and the few-shot learning models.
Furthermore, our current experiments are  a simulate of the user-defined scenario.
In the future we will conduct more  experiments which resemble more realistic situations such as mixing keywords with different languages.

\vspace{4mm}

\newpage
\bibliographystyle{IEEEtran}

\bibliography{mybib}

% \begin{thebibliography}{9}
% \bibitem[1]{Davis80-COP}
%   S.\ B.\ Davis and P.\ Mermelstein,
%   ``Comparison of parametric representation for monosyllabic word recognition in continuously spoken sentences,''
%   \textit{IEEE Transactions on Acoustics, Speech and Signal Processing}, vol.~28, no.~4, pp.~357--366, 1980.
% \bibitem[2]{Rabiner89-ATO}
%   L.\ R.\ Rabiner,
%   ``A tutorial on hidden Markov models and selected applications in speech recognition,''
%   \textit{Proceedings of the IEEE}, vol.~77, no.~2, pp.~257-286, 1989.
% \bibitem[3]{Hastie09-TEO}
%   T.\ Hastie, R.\ Tibshirani, and J.\ Friedman,
%   \textit{The Elements of Statistical Learning -- Data Mining, Inference, and Prediction}.
%   New York: Springer, 2009.
% \bibitem[4]{YourName17-XXX}
%   F.\ Lastname1, F.\ Lastname2, and F.\ Lastname3,
%   ``Title of your INTERSPEECH 2020 publication,''
%   in \textit{Interspeech 2020 -- 20\textsuperscript{th} Annual Conference of the International Speech Communication Association, September 15-19, Graz, Austria, Proceedings, Proceedings}, 2020, pp.~100--104.
% \end{thebibliography}

\end{document}